\documentclass[letterpaper]{article} 
\usepackage{aaai2027}
\usepackage[hyphens]{url}  
\usepackage{graphicx} 
\usepackage{natbib}  
\usepackage{caption} 
\usepackage{algorithm}
\usepackage{algorithmic}

\usepackage{multirow}
\usepackage{amsmath}
\usepackage{amssymb}
\usepackage{xcolor}

\usepackage{newfloat}
\usepackage{listings}
\DeclareCaptionStyle{ruled}{labelfont=normalfont,labelsep=colon,strut=off} 
\floatstyle{ruled}
\newfloat{listing}{tb}{lst}{}
\floatname{listing}{Listing}

\usepackage{booktabs}

\nocopyright

\title{Predicting Functions, Not Features: KANs with Function-Space Joint-Embedding Predictive Learning for Medical Image Segmentation}
\author{
    Yungeng Liu\textsuperscript{\rm 1}\equalcontrib,
    Xuanzi Fang\textsuperscript{\rm 1}\equalcontrib,
    Yuge Zhang\textsuperscript{\rm 2},
    Shuqi Ren\textsuperscript{\rm 1},
    Haijin Zeng\textsuperscript{\rm 1},
    Yongyong Chen\textsuperscript{\rm 1}\corresponding
}
\affiliations{
    \textsuperscript{\rm 1}Harbin Institute of Technology,Shenzhen\\
    \textsuperscript{\rm 2}The University of Hong Kong\\
    25B951006@stu.hit.edu.cn, firetreehouse0@gmail.com, 
    u3645972@connect.hku.hk,
    konosubame@gmail.com,
    haijin.zeng2018@gmail.com,
    cyy2020@hit.edu.cn
}

\begin{document}

\maketitle

\begin{abstract}

Kolmogorov--Arnold Networks (KANs) introduce explicit functional representations by parameterizing each network edge as a learnable univariate function. However, existing KAN-based segmentation models optimize edge functions only through objectives defined after edge aggregation, leaving individual functions without an explicit pre-aggregation learning target.
To address this limitation, we propose Function-Space Joint-Embedding Predictive Learning (FS-JEPA) for medical image segmentation. Our FS-JEPA framework moves predictive learning into the pre-aggregation function space of KANs. A masked online branch predicts structured signatures of sampled KAN edge functions generated by a full-context exponential moving average target branch, while shared edge indices preserve correspondence between predictions and targets.
Rather than predicting an isolated edge response, we represent each sampled edge function using a multi-radius signature composed of function evaluations around its input anchor. This structured representation captures local functional variations that cannot be characterized by a single response and provides a more informative predictive target. The function-space objective is jointly optimized with the segmentation loss during training, while the predictive branch is removed at inference. Experiments on five medical image segmentation benchmarks show that our FS-JEPA achieves the best average Dice and outperforms the strongest competing KAN-based method by +2.25 percentage points.
\end{abstract}

\section{Introduction}
\label{sec:introduction}

\begin{figure}[t]
\centering
\includegraphics[width=1\linewidth]{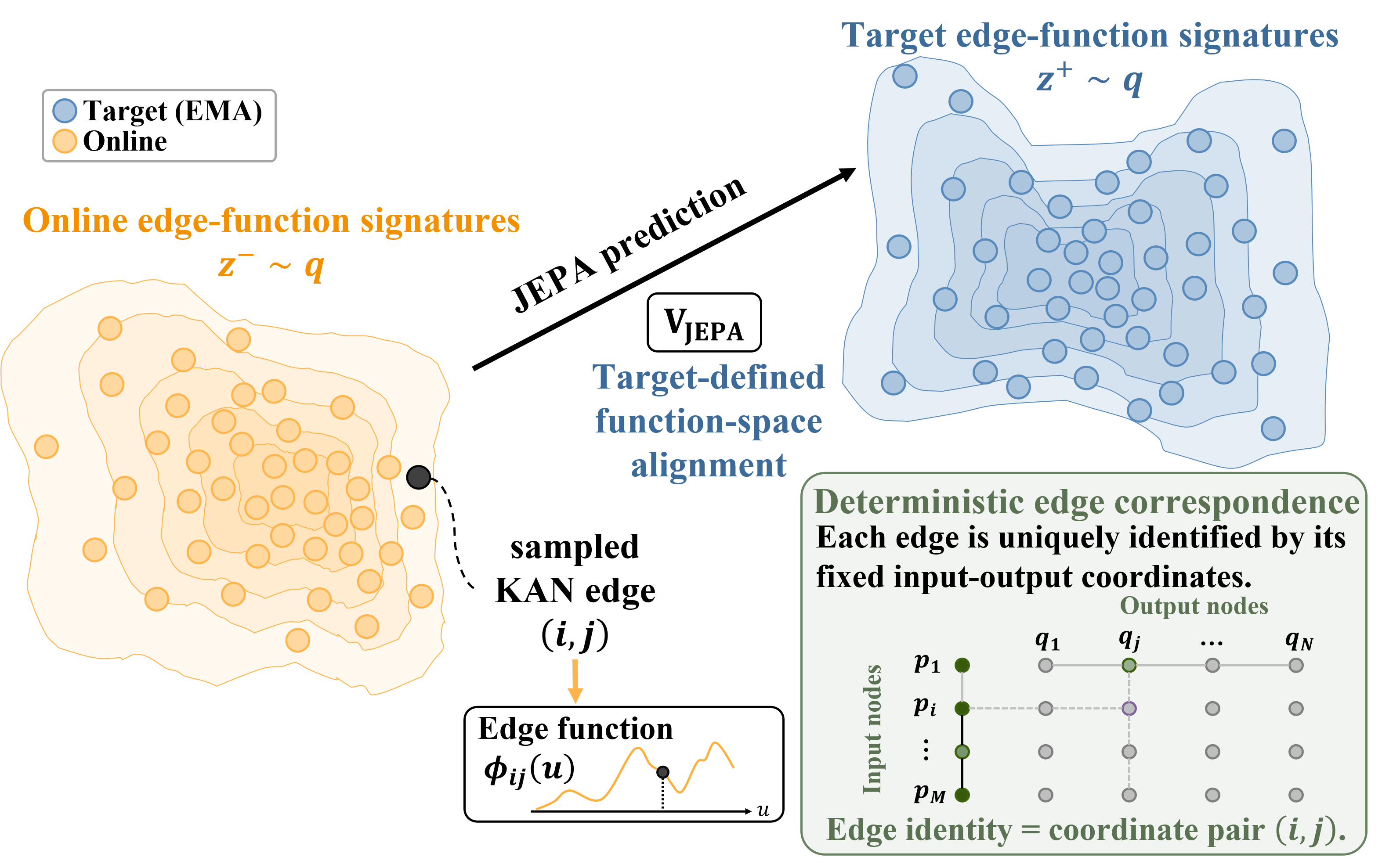}
\caption{\textbf{Motivation of Our FS-JEPA for KANs.}
To overcome the limitations of feature-level prediction, predictive learning is extended to structured representations of pre-aggregation KAN edge functions. Multi-radius signatures characterize local functional variations, enabling predictive learning directly in KAN function space.}
\label{fig:motivation}
\end{figure}

\begin{figure*}[ht]
    \centering
    \includegraphics[width=0.9\linewidth]{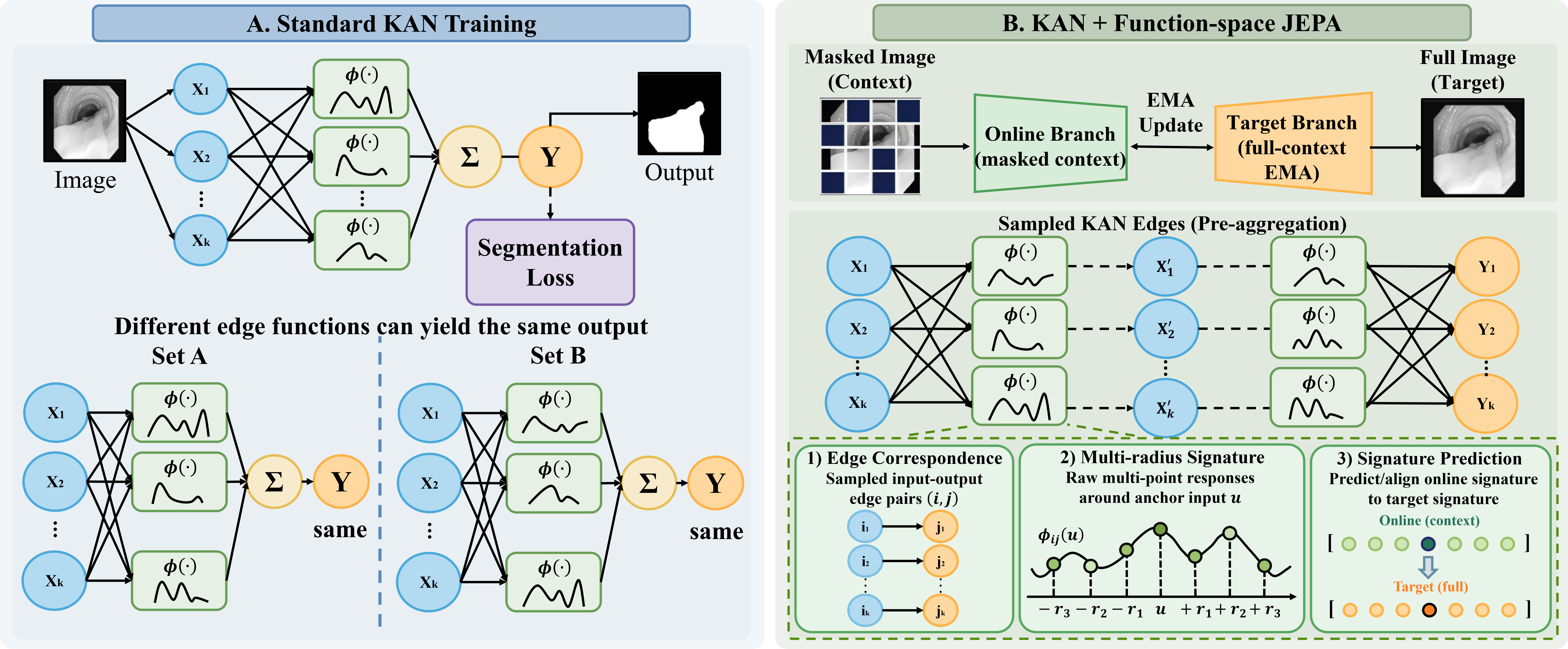}
    \caption{\textbf{Standard KAN training versus the proposed FS-JEPA framework.} (A) Standard KAN training optimizes edge functions only through post-aggregation task supervision, under which different edge-function configurations may yield the same node output. (B) Our framework directly performs predictive learning on sampled pre-aggregation edge functions by predicting their multi-radius signatures from masked online contexts to full-context EMA targets, with shared edge indices ensuring correspondence.}
    \label{fig:versus}
\end{figure*}

Medical image segmentation requires dense prediction under weak contrast, ambiguous boundaries, artifacts, and large appearance variations. Encoder--decoder architectures address these challenges through multi-scale feature fusion and improved token interaction \cite{ronneberger2015unet,zhou2018unetpp,valanarasu2022unext,liu2024rolling}, but their nonlinear transformations remain implicitly defined by scalar weights and fixed activations rather than explicit functional objects.

Kolmogorov--Arnold Networks (KANs) place a learnable univariate function on each network edge \cite{liu2025kan}. Given an input $x_i$, the $j$-th output is defined as $y_j=\sum_i\phi_{j,i}(x_i)$, where $\phi_{j,i}$ denotes the function from input node $i$ to output node $j$. This representation has been introduced into medical image segmentation by U-KAN \cite{li2025ukan} and extended to visual token processing by KAT \cite{yang2025kat}. However, existing KAN-based vision models are optimized only through losses on aggregated node outputs. Such supervision constrains the sum of incoming edge responses, allowing different edge-function configurations to produce similar outputs and leaving individual functions without an explicit pre-aggregation objective. Moreover, each function is evaluated only at an input-dependent anchor during a forward pass, although functions with the same response at that point may behave differently nearby. Thus, edge-specific behavior and local functional variation remain only implicitly constrained by the downstream task.

Joint-embedding predictive learning provides a way to learn structured
representations beyond task supervision. Instead of reconstructing raw
inputs, JEPA predicts target representations from contextual observations
\cite{assran2023ijepa}. Existing methods, however, mainly operate on
image-, token-, or feature-level embeddings and do not consider learnable
neural functions as prediction targets. As shown in Fig.~\ref{fig:motivation}, extending predictive learning to
KAN function space is non-trivial. First, a single edge response cannot
characterize the local behavior of a function. Second, because different
KAN edges represent different input--output connections, the sampled edges
in the online and target branches must remain explicitly aligned. These
challenges raise the question: \emph{can predictive learning be extended
from feature representations to structured representations of learnable
neural functions?}

We answer this question with FS-JEPA for KAN-based medical image segmentation. As illustrated in Fig.~\ref{fig:versus}, during training, a masked online branch predicts structured signatures generated
by a full-context exponential moving average target branch before KAN edge
aggregation. The two branches share the same sampled edge indices, ensuring
that each prediction corresponds to the same input--output edge function.
Rather than predicting an isolated response, we represent each sampled edge
function using a multi-radius signature composed of function evaluations
around its input anchor. This signature captures local functional variations
and provides an explicit pre-aggregation predictive objective for individual
KAN edges. The objective is jointly optimized with the supervised
segmentation loss, while all predictive components are removed at inference.
Across five medical image segmentation benchmarks, FS-JEPA achieves the
highest average Dice and outperforms the strongest competing KAN-based
method by +2.25 percentage points.
Our contributions are summarized as follows:
\begin{itemize}
\item We formulate joint-embedding predictive learning in the native pre-aggregation function space of KANs, shifting the prediction target from feature embeddings
to structured edge-function signatures.

\item We introduce a multi-radius function signature representation that captures local functional behavior of KAN edge functions and provides an effective prediction target for function-space learning.

\item We develop a training-only predictive framework that jointly learns KAN edge-function signatures and segmentation representations from random initialization while maintaining online-target edge correspondence under controlled training cost.
\end{itemize}

\section{Related Work}
\label{sec:related-work}

\subsection{Medical Image Segmentation}

U-Net established the encoder--decoder paradigm with skip connections for
medical image segmentation and inspired numerous variants
\cite{ronneberger2015unet,wang2025comprehensive,xia2025comprehensive}.
UNet++ reduces semantic gaps through nested skip pathways
\cite{zhou2018unetpp}, while recent architectures explore efficient feature
interaction, such as tokenized MLPs in UNeXt, compact long-range modeling in
Rolling-UNet, and boundary-aware attention in PBE-UNet
\cite{valanarasu2022unext,liu2024rolling,wang2026pbe}. Beyond architectural
design, optimization strategies including deep supervision and boundary-aware
objectives are explored
\cite{lee2015deeply,kervadec2019boundary}. 
\textit{These techniques are adopted as a fixed segmentation scaffold, while our contribution focuses on learning structured representations of KAN functions.}

\subsection{Kolmogorov--Arnold Networks for Vision}

KANs replace fixed activations with learnable univariate edge functions,
providing an explicit functional representation beyond conventional
weight-based networks \cite{liu2025kan}. This formulation has inspired various
vision architectures, including U-KAN for medical segmentation, KAT for visual
token processing, MM-UKAN++ for boundary refinement, and UUEKAN with
uncertainty-guided attention \cite{li2025ukan,yang2025kat,MMUKANpp,CHEN2026110649}.
\textit{However, existing KAN-based vision models still optimize edge functions indirectly through task objectives after edge aggregation. Although gradients are propagated to individual edges, the underlying functional structures are not explicitly modeled as learning targets. Our work addresses this limitation by introducing predictive learning directly on pre-aggregation KAN edge functions.}

\subsection{Joint-Embedding Predictive Learning}

Self-supervised predictive learning aims to learn representations by predicting
intrinsic data structures rather than relying on manual annotations. BYOL and
VICReg demonstrate the effectiveness of prediction-based objectives without
contrastive negative samples \cite{grill2020byol,bardes2022vicreg}.
JEPA further shifts prediction from input reconstruction to latent
representation prediction, where I-JEPA learns target representations from
visible contexts and subsequent variants extend this paradigm to temporal and
video domains \cite{assran2023ijepa,monemi2025tutorial,bardes2023mc,bardes2024revisiting}.
\textit{However, existing JEPA methods remain limited to generic feature embeddings and do not consider learnable neural operators as prediction targets. We extend JEPA from feature space to KAN function space, where edge functions are represented by structured multi-radius signatures and directly optimized through predictive learning.}

\begin{figure*}[th]
    \centering
    \includegraphics[width=1\linewidth]{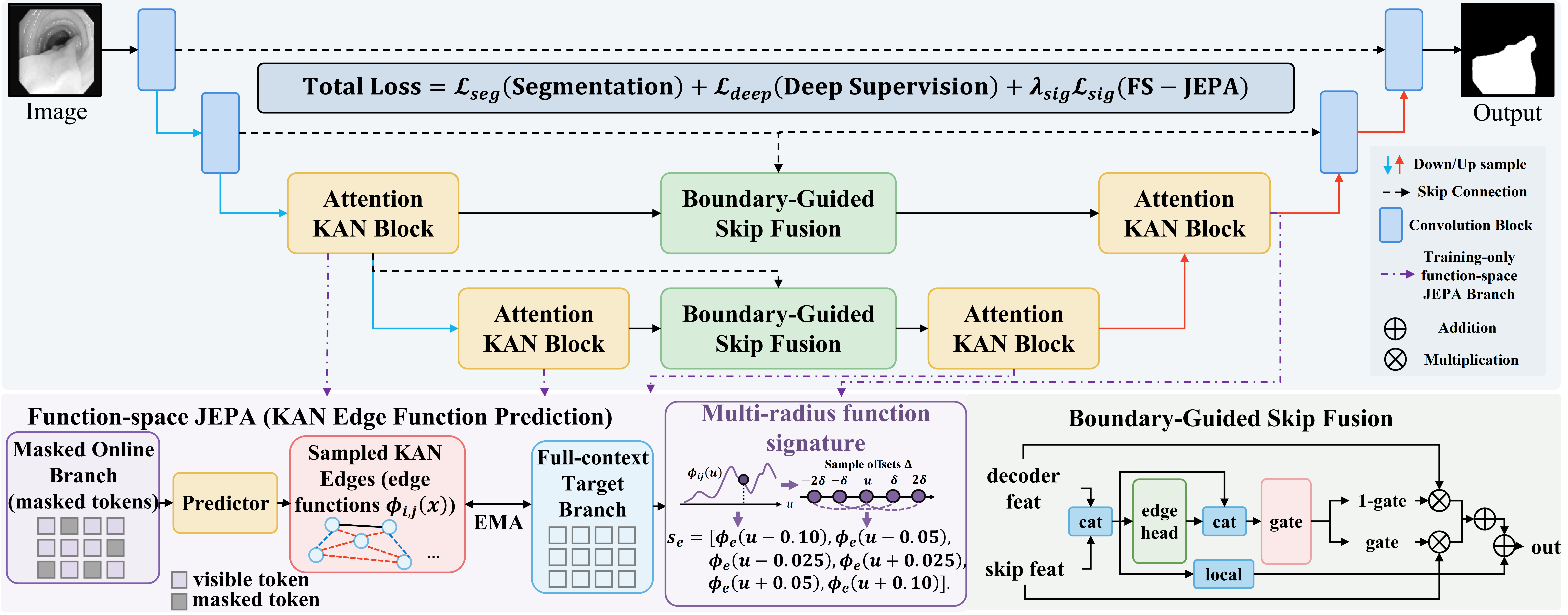}
    \caption{\textbf{Overview of the FS-JEPA framework.} The segmentation network is jointly optimized with a training-only predictive branch that learns multi-radius signatures of sampled KAN edge functions before aggregation. The online branch predicts EMA target signatures under explicit edge correspondence, enabling predictive representation learning in KAN function space.}
    \label{fig:framework}
\end{figure*}

\section{Method}
\label{sec:method}

\subsection{Overview}

Let $f_\theta$ denote a U-shaped KAN segmentation network. The segmentation
stream receives the complete image and is optimized with standard supervision. As illustrated in Fig.~\ref{fig:framework}, our framework adopts established components, including Attention KAN Blocks and Boundary-Guided Skip Fusion. The Attention KAN Block follows recent attention-enhanced KAN designs such as KAT ~\cite{yang2025kat}, while the boundary-guided fusion module is used to improve feature integration across scales. Full module definitions and layer-wise configurations are provided in Section A of our Supplementary. 
During training, we attach a function-space JEPA branch to the KAN blocks. The
online branch receives a partially masked token sequence and predicts function
signatures generated by a full-context EMA target branch, following the
predictive learning principle of JEPA.
Unlike feature-level predictive learning, our objective is defined before KAN
edge aggregation and directly targets individual edge functions
$\phi_{j,i}$. Since the complete target representation is generated from the
same sampled KAN edges, the prediction preserves edge-level correspondence.
Instead of predicting isolated edge responses, we represent each function using
a multi-radius signature that captures local functional behavior. This design
enables direct predictive learning over structured KAN edge functions without contrastive negatives, reconstruction losses, or manually imposed derivative constraints.

\subsection{KAN Edge Functions and Edge Sampling}
\label{sec:selected-edge}

\begin{table*}[t]

\centering
\small

\setlength{\tabcolsep}{2pt}

\begin{tabular*}{\textwidth}{@{\extracolsep{\fill}}lcccccccccccccc}
\toprule

\multirow{2}{*}{\textbf{Methods}}
& \multicolumn{2}{c}{\textbf{BUSI}}
& \multicolumn{2}{c}{\textbf{DDTI}}
& \multicolumn{2}{c}{\textbf{TN3K}}
& \multicolumn{2}{c}{\textbf{CVC}}
& \multicolumn{2}{c}{\textbf{GlaS}}
& \multicolumn{2}{c}{\textbf{Average}}
& \textbf{Params(M)}
\\

\cmidrule(lr){2-3}
\cmidrule(lr){4-5}
\cmidrule(lr){6-7}
\cmidrule(lr){8-9}
\cmidrule(lr){10-11}
\cmidrule(lr){12-13}

& Dice & IoU
& Dice & IoU
& Dice & IoU
& Dice & IoU
& Dice & IoU
& Dice & IoU
& \\

\midrule


UNet++\shortcite{zhou2018unetpp}
&0.7093&0.6232
&0.7179&0.6030
&0.8008&0.7137
&\underline{0.8702}&\underline{0.8045}
&\underline{0.9105}&\underline{0.8423}
&0.8017&0.7173
&9.16
\\

UNeXt\shortcite{valanarasu2022unext}
&0.6671&0.5655
&0.6657&0.5325
&0.7593&0.6582
&0.7262&0.6211
&0.8469&0.7427
&0.7330&0.6240
&1.47
\\

Rolling-UNet-S\shortcite{liu2024rolling}
&0.6932&0.5967
&0.7077&0.5826
&0.7985&0.7077
&0.7978&0.7117
&0.8903&0.8097
&0.7775&0.6817
&1.78
\\

PBE-UNet\shortcite{wang2026pbe}
&0.7490&0.6574
&0.6912&0.5664
&0.8065&0.7167
&0.8268&0.7478
&0.8264&0.7198
&0.7800&0.6816
&4.26
\\

\midrule

\multicolumn{14}{c}{\textit{KAN-based Methods}}
\\
\midrule

U-KAN\shortcite{li2025ukan}
&0.7011&0.6097
&0.6893&0.5610
&0.7865&0.6948
&0.7607&0.6741
&0.8680&0.7748
&0.7611&0.6629
&6.36
\\

MM-UKAN++\shortcite{MMUKANpp}
&0.5786&0.4808
&0.6187&0.4802
&0.7371&0.6370
&0.7065&0.5946
&0.7826&0.6541
&0.6847&0.5693
&7.68
\\

KMUNet\shortcite{zhang2025kmunet}
&0.7494&0.6599
&0.7372&\underline{0.6287}
&{0.8132}&{0.7237}
&0.8191&0.7431
&0.8982&0.8239
&0.8034&0.7159
&7.01
\\

Implicit U-KAN2.0\shortcite{cheng2025implicit}
&\underline{0.7644}&0.6722
&0.7067&0.5854
&0.7699&0.6685
&0.7962&0.7145
&0.8718&0.7855
&0.7818&0.6852
&7.85
\\

UUEKAN\shortcite{CHEN2026110649}
&0.7569&\underline{0.6732}
&\textbf{0.7447}&\textbf{0.6338}
&\underline{0.8206}&\underline{0.7349}
&0.8346&0.7697
&0.8992&0.8253
&\underline{0.8112}&\underline{0.7274}
&35.02
\\

\midrule

Matched Scaffold
&0.7808&0.7017
&0.7289&0.6152
&0.8217&0.7332
&0.8798&0.8113
&0.9149&0.8505
&0.8252&0.7424
&7.74
\\

\textbf{FS-JEPA(Ours)}
&\textbf{0.7926}&\textbf{0.7035}
&\underline{0.7432}&0.6275
&\textbf{0.8258}&\textbf{0.7401}
&\textbf{0.8852}&\textbf{0.8199}
&\textbf{0.9215}&\textbf{0.8611}
&\textbf{0.8337}&\textbf{0.7504}
&7.74
\\

\bottomrule
\end{tabular*}
\caption{Comparison with state-of-the-art methods on five medical image segmentation benchmarks. The best results are highlighted in \textbf{bold} and the second-best results are \underline{underlined}. All results are reported using image-level Dice and IoU. The Matched Scaffold denotes the complete segmentation architecture in
Fig.~\ref{fig:framework} trained without the FS-JEPA branch.}
\label{tab:main-results}
\end{table*}

\begin{figure*}[t]
    \centering
    \includegraphics[width=0.9\linewidth]{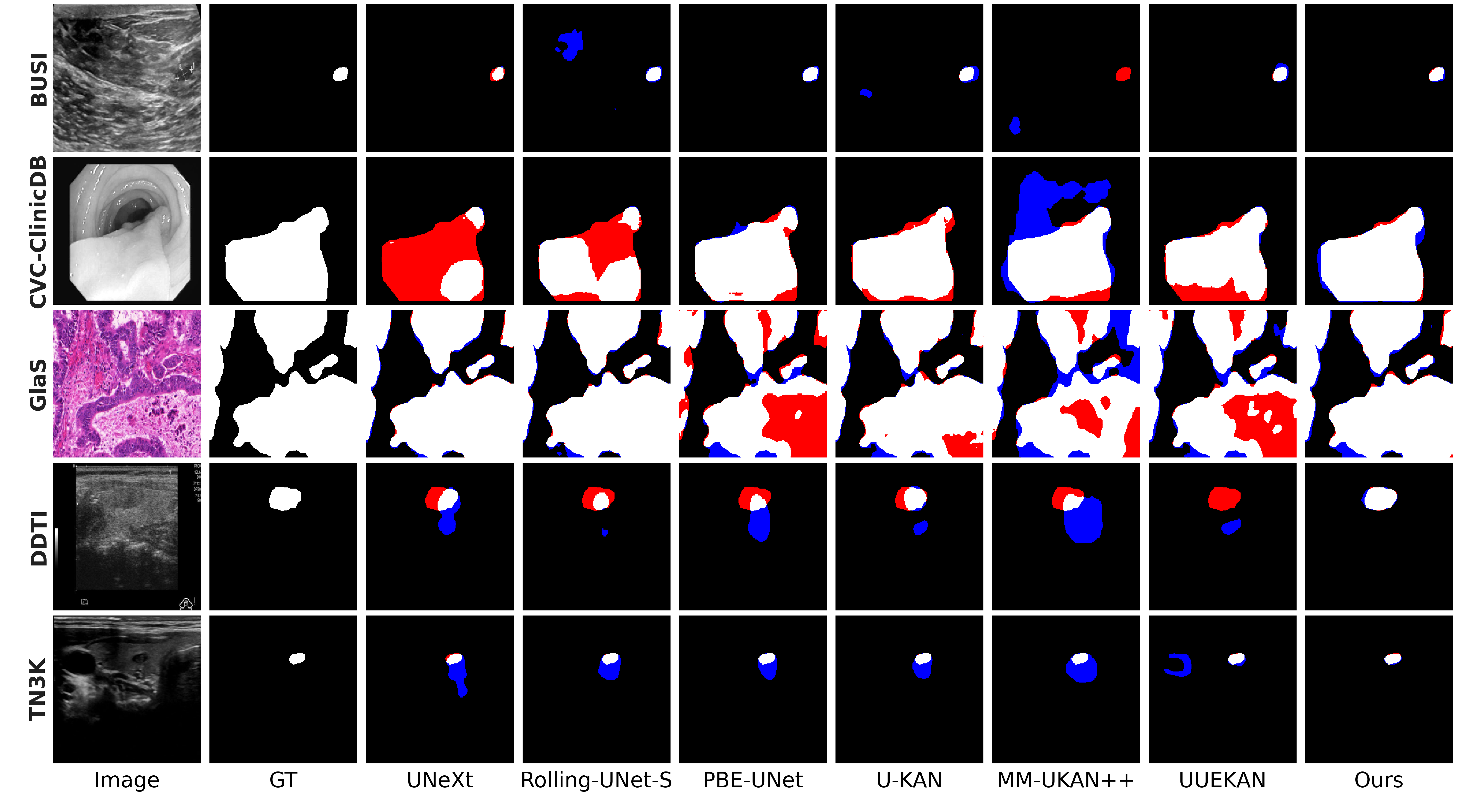}
    \caption{
    Qualitative comparison of segmentation results. From left to right: input
    images, ground truth, and predictions from different methods. White, red, and
    blue regions indicate true positive, false positive, and false negative
    regions, respectively.
    }
    \label{fig:sixdataset}
\end{figure*}

KANs represent network transformations through learnable functions assigned to individual edges. Consider a KAN layer with  $d_{\mathrm{in}}$ input nodes and $d_{\mathrm{out}}$ output nodes. In the spline-based parameterization adopted in our backbone, the response of edge $(i,j)$ to a scalar input $u$ is
\begin{equation}
    \phi_{j,i}(u)
    =
    w^{b}_{j,i}\operatorname{SiLU}(u)
    +
    s_{j,i}
    \sum_{\ell=1}^{L}
    w^{s}_{j,i,\ell}B^{p}_{\ell}(u),
\end{equation}
where $w^{b}_{j,i}$ denotes the base function coefficient,
$w^{s}_{j,i,\ell}$ denotes the spline coefficients, $s_{j,i}$ is the
learnable spline scale, and $B^{p}_{\ell}$ represents a B-spline basis function
of order $p$.
The standard KAN forward operation aggregates all incoming edge functions:$y_j=\sum_i\phi_{j,i}(x_i).$
Although this aggregation enables efficient computation, it also introduces a
many-to-one mapping from edge functions to node representations. Different
combinations of edge functions may produce similar aggregated outputs, making
the individual functional behaviors of edges difficult to supervise through
task objectives alone. Therefore, our framework exposes the pre-aggregation
edge functions as predictive targets.
Directly constructing responses for all possible edges requires
$O(BNd_{\mathrm{in}}d_{\mathrm{out}})$ auxiliary activations, which becomes
prohibitively expensive for large KAN layers. Instead, we randomly sample
$K$ input-output edge pairs:$\mathcal{S}=\{(i_k,j_k)\}_{k=1}^{K}.$
For sampled edges, we define a selected-edge operator:
\begin{equation}
    \Phi_{\mathcal{S}}(U)_{b,n,k}
    =
    \phi_{j_k,i_k}(U_{b,n,i_k}),
\end{equation}
which evaluates only the required edge functions by gathering their
corresponding spline parameters and knot grids. This operation is equivalent
to computing the complete edge-response tensor followed by selecting the same
entries, while reducing auxiliary activation cost from
$O(BNd_{\mathrm{in}}d_{\mathrm{out}})$ to $O(BNK)$.
The sampled edge functions obtained by $\Phi_{\mathcal{S}}$ are subsequently
used to construct multi-radius function signatures for the proposed
function-space predictive objective.

\subsection{Multi-radius Edge Function Signature}
\label{sec:function-signature}

A single point evaluation cannot fully characterize a KAN edge function. Two
different functions may produce identical responses at one input location while
exhibiting different local behaviors around that point. Therefore, we represent
each sampled edge function using a multi-radius signature constructed from
neighboring function evaluations.

For a sampled edge $e=(i,j)$ with anchor input $u$, we define a set of local
offsets:
\begin{equation}
    \mathcal{D}
    =
    \{-0.10,-0.05,-0.025,0.025,0.05,0.10\}.
\end{equation}

The function signature is obtained by evaluating the same edge function at multiple neighboring locations:
\begin{equation}
    s_e(\Delta)
    =
    \phi_e(u+\Delta),
    \quad
    \Delta\in\mathcal{D}.
\end{equation}

The final multi-radius signature is defined as:
\begin{equation}
\begin{aligned}
s_e =
[
&\phi_e(u-0.10),
\phi_e(u-0.05),
\phi_e(u-0.025),\\
&\phi_e(u+0.025),
\phi_e(u+0.05),
\phi_e(u+0.10)
].
\end{aligned}
\end{equation}

Unlike single-response prediction, the proposed signature transforms each KAN
edge function into a structured local representation. By preserving the
response pattern around the input anchor, it captures local functional
variations and provides a more informative prediction target for
function-space JEPA without introducing manually designed auxiliary losses.

\subsection{Deterministic Edge Correspondence}
\label{sec:edge-coordinate}

The stochastic sampling of KAN edges introduces a correspondence problem: the
same sampled position may refer to different input-output edges across
iterations. Since function signatures describe local functional behavior but
do not inherently encode edge identity, explicit edge correspondence is
required for stable predictive learning.
For a sampled edge $(i,j)$, we encode its source and target node positions using
deterministic coordinates:
\begin{equation}
    p_i=2\frac{i-1}{d_{\mathrm{in}}-1}-1,
    \qquad
    p_j=2\frac{j-1}{d_{\mathrm{out}}-1}-1 .
\end{equation}
The coordinate pair $(p_i,p_j)$ is concatenated with the multi-radius function
signature and provided as the condition of the signature predictor. Since the
online and target branches share identical sampled edge indices, these
deterministic coordinates establish consistent correspondence between predicted
and target signatures without introducing additional learnable parameters.
This design separates functional representation from edge indexing: the
signature captures the local behavior of an edge function, while the
coordinates identify the corresponding input-output connection.

\begin{figure*}[h]
    \centering
    \includegraphics[width=1\textwidth]{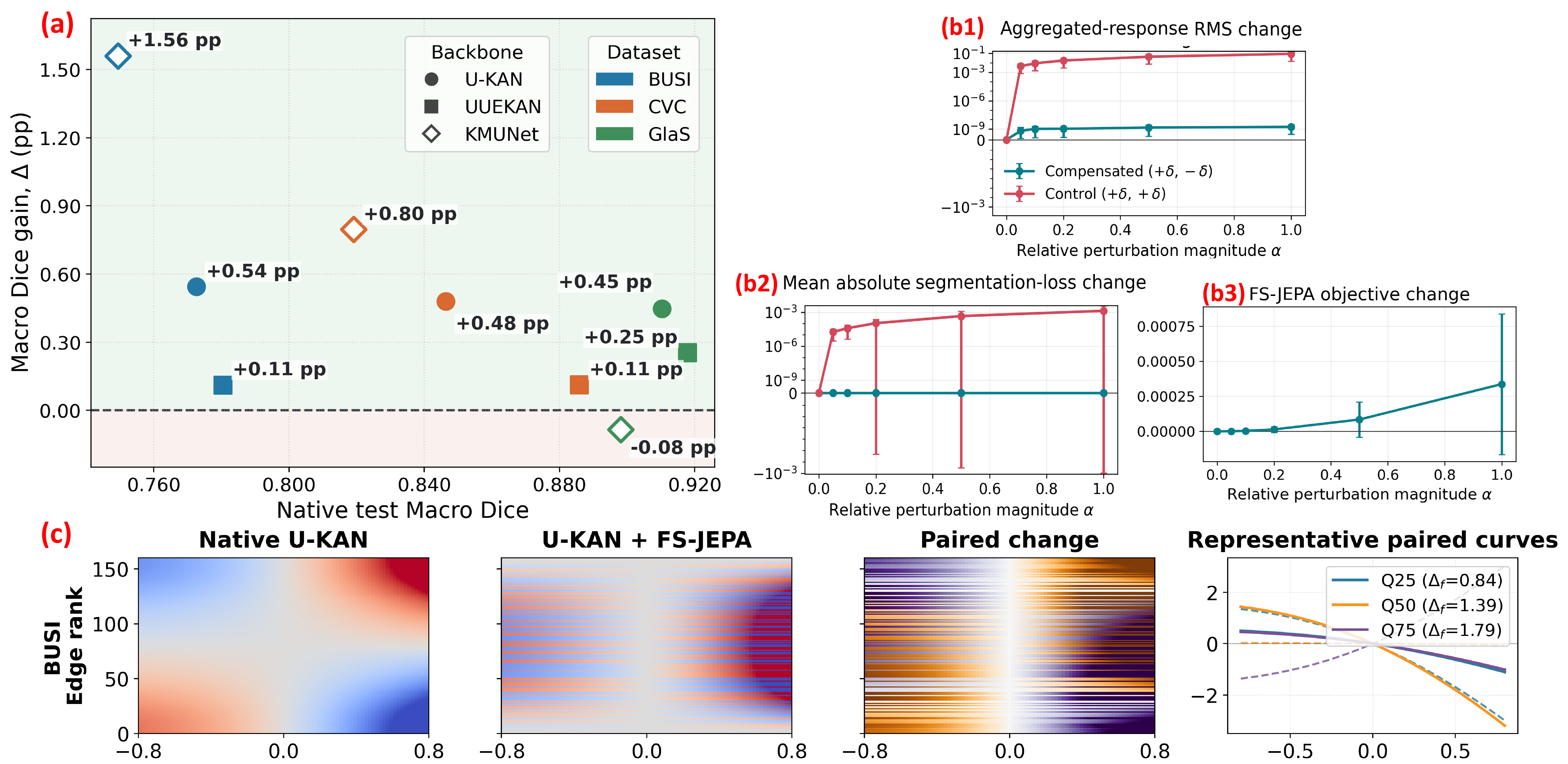}
\caption{
Mechanistic and cross-backbone analysis of FS-JEPA.
(a) Paired Dice gains obtained by applying FS-JEPA to U-KAN,
UUEKAN, and KMUNet across BUSI, CVC, and GlaS.
(b) Sensitivity of the aggregated KAN response, segmentation loss,
and FS-JEPA objective to compensated and control edge perturbations.
(c) Paired edge-function analysis on BUSI, showing the responses of
native U-KAN and U-KAN with FS-JEPA, their edge-wise changes, and
representative paired curves.
}
    \label{fig:mechanistic_analysis}
\end{figure*}

\subsection{Function Signature Prediction Objective}
\label{sec:signature-objective}

Given the predicted signature $\hat{s}_e$ from the masked online branch and the target signature $s_e^+$ generated by the full-context EMA branch, we formulate the function-space predictive objective by aligning the two signatures.
Specifically, the online predictor receives the masked context representation together with the sampled edge information and produces: $\hat{s}_e=q_{\psi}(z_e),
s_e^+=\operatorname{sg}(s_e^{EMA}),$ Here, $z_e$ concatenates the masked online edge response, its input anchor, and the deterministic coordinates $(p_i,p_j)$ of edge $e=(i,j)$, and $q_{\psi}$ denotes the signature predictor and
$\operatorname{sg}(\cdot)$ indicates stop-gradient operation. The EMA target
branch provides a stable prediction target without receiving gradient updates.
The predictive loss is defined as: $\mathcal{L}_{sig}=\operatorname{SmoothL1}(\hat{s}_e,s_e^+).$ 
Following the JEPA principle, the online branch learns to infer the complete
function signature from partially observed contexts, while the EMA branch
provides a slowly evolving target representation. Unlike feature-level
predictive learning, the prediction target here is defined over the structured
representation of individual KAN edge functions before aggregation. Therefore,
the objective directly constrains the functional behavior of KAN edges rather
than only their aggregated node outputs.
Since the online and target branches share identical sampled edge indices, each
predicted signature corresponds to the same input-output edge function. The
deterministic edge coordinates provide explicit correspondence information,
while the multi-radius signature captures local functional behavior around the
input anchor.
The proposed objective learns structured KAN edge-function representations
without auxiliary constraints or contrastive objectives. By predicting
multi-radius function signatures, it enables predictive learning directly in
KAN function space beyond conventional feature representations.

\subsection{Joint Optimization and Inference}
\label{sec:joint-objective}

The proposed function-space predictive objective is jointly optimized with the
supervised segmentation task. The segmentation objective consists of the main
segmentation loss $\mathcal{L}_{seg}$ and decoder-side deep supervision loss
$\mathcal{L}_{deep}$. For the four KAN blocks equipped with the predictive branch, the final training objective is defined as:
\begin{equation}
    \mathcal{L}_{total}
    =
    \mathcal{L}_{seg}
    +
    \lambda_{deep}\mathcal{L}_{deep}
    +
    \frac{\lambda_{sig}}{4}
    \sum_{b=1}^{4}
    \mathcal{L}_{sig}^{(b)} .
\end{equation}
The segmentation and function-space predictive objectives are jointly optimized from random initialization, without pretrained checkpoints or external teacher models. The EMA target branch and signature predictor are used only during training and removed at inference, leaving the original segmentation pathway unchanged.

\begin{table*}[t]
    \centering
    \small
    \setlength{\tabcolsep}{7pt}
    \begin{tabular}{llccc}
        \toprule
        Predictive method & Target space & Pre-sum
        & Dice $\uparrow$ & IoU $\uparrow$ \\
        \midrule
        No predictive objective & -- & --
        & $0.7808\!\pm\!0.0093$ & $0.7017\!\pm\!0.0103$ \\
        Token-JEPA & Token feature & --
        & $0.7808\!\pm\!0.0097$ & $0.6921\!\pm\!0.0114$ \\
        Node-JEPA & KAN node feature & --
        & $0.7845\!\pm\!0.0145$ & $0.6995\!\pm\!0.0151$ \\
        Edge-JEPA & Single edge response & \checkmark
        & $0.7771\!\pm\!0.0123$ & $0.6933\!\pm\!0.0109$ \\
        \textbf{Ours} & \textbf{Multi-radius edge-function signature}
        & \checkmark
        & $\mathbf{0.7926\!\pm\!0.0049}$
        & $\mathbf{0.7035\!\pm\!0.0054}$ \\
        \bottomrule
    \end{tabular}
    \caption{Comparison of predictive target spaces on BUSI under the same segmentation scaffold and 300-epoch budget. ``Pre-sum'' indicates that prediction is performed before KAN edge aggregation.}
    \label{tab:target-space-ablation}
\end{table*}


\begin{table*}[t]
    \centering
    \small
    \begingroup
    \setlength{\tabcolsep}{2pt}
    \renewcommand{\arraystretch}{1.0}
    \begin{tabular}{lcccccc}
        \toprule
        Configuration
        & Multi-radius
        & Rel. MAE $\downarrow$
        & NRMSE $\downarrow$
        & Pearson $r$ $\uparrow$
        & Std. ratio $\rightarrow 1$
        & Corr. error $\downarrow$ \\
        \midrule

        Response only
        & --
        & $0.9910\!\pm\!0.0083$
        & $0.9893\!\pm\!0.0124$
        & $0.3139\!\pm\!0.1191$
        & $0.0576\!\pm\!0.0746$
        & $0.0859\!\pm\!0.0063$ \\

        Raw multi-point
        & \checkmark
        & $\mathbf{0.9175\!\pm\!0.0449}$
        & $\mathbf{0.8999\!\pm\!0.0429}$
        & $\mathbf{0.4622\!\pm\!0.0796}$
        & $\mathbf{0.2974\!\pm\!0.0910}$
        & $0.0679\!\pm\!0.0029$ \\

        Centered signature
        & \checkmark
        & $1.0052\!\pm\!0.0023$
        & $0.9923\!\pm\!0.0038$
        & $0.1490\!\pm\!0.0355$
        & $0.0637\!\pm\!0.0178$
        & $0.0586\!\pm\!0.0084$ \\

        Diagonal signature
        & \checkmark
        & $1.0052\!\pm\!0.0009$
        & $0.9921\!\pm\!0.0001$
        & $0.1608\!\pm\!0.0062$
        & $0.0603\!\pm\!0.0028$
        & $\mathbf{0.0542\!\pm\!0.0058}$ \\

        ZCA signature
        & \checkmark
        & $0.9928\!\pm\!0.0019$
        & $0.9949\!\pm\!0.0029$
        & $0.2127\!\pm\!0.0531$
        & $0.0244\!\pm\!0.0087$
        & $0.0599\!\pm\!0.0046$ \\
        \bottomrule
    \end{tabular}
    \endgroup
    \caption{Function-prediction diagnostics for different edge-function
    signature constructions on BUSI. The matched scaffold has no predictive branch and
    therefore no function-prediction diagnostics.}
    \label{tab:function-signature-diagnostics}
\end{table*}

\begin{table}[t]
    \centering
    \small
    \begingroup
    \setlength{\tabcolsep}{2pt}
    \renewcommand{\arraystretch}{1.0}
    \begin{tabular}{lccc}
        \toprule
        Configuration
        & \shortstack{Dice $\uparrow$}
        & \shortstack{IoU $\uparrow$}
        & $\Delta$Dice \\
        \midrule
        \shortstack[l]{Matched scaffold}
        & $0.7808\!\pm\!0.0093$
        & $0.7017\!\pm\!0.0103$
        & $+0.0037$ \\

        \shortstack[l]{Response only}
        & $0.7771\!\pm\!0.0123$
        & $0.6933\!\pm\!0.0109$
        & $0.0000$ \\

        \shortstack[l]{Raw multi-point}
        & $\mathbf{0.7926\!\pm\!0.0049}$
        & $0.7035\!\pm\!0.0054$
        & $\mathbf{+0.0155}$ \\

        \shortstack[l]{Centered signature}
        & $0.7786\!\pm\!0.0025$
        & $0.6925\!\pm\!0.0039$
        & $+0.0015$ \\

        \shortstack[l]{Diagonal signature}
        & $0.7897\!\pm\!0.0167$
        & $\mathbf{0.7047\!\pm\!0.0132}$
        & $+0.0126$ \\

        \shortstack[l]{ZCA signature}
        & $0.7810\!\pm\!0.0048$
        & $0.6927\!\pm\!0.0066$
        & $+0.0039$ \\
        \bottomrule
    \end{tabular}
    \endgroup
    \caption{Segmentation performance of edge-function signature variants on BUSI. }
    \label{tab:function-signature-segmentation}
\end{table}

\section{Experiments}
\label{sec:experiments}

\textbf{Experimental Setup:}
To evaluate the proposed method, as shown in Table~\ref{tab:main-results},
we conduct experiments on five medical image segmentation benchmarks covering different imaging modalities: BUSI~\cite{aldhabyani2020busi} and DDTI~\cite{pedraza2015ddti}
for ultrasound-based lesion and nodule segmentation, TN3K~\cite{gong2023trfe} for thyroid region and nodule segmentation, CVC-ClinicDB~\cite{bernal2015wmdova}
for polyp segmentation, and GlaS~\cite{sirinukunwattana2017glas} for gland segmentation in histopathology images. All methods are evaluated under the same data partition for each dataset to ensure a fair comparison. 
All models are trained from random initialization under identical data splits and optimization settings without external pre-training or teacher models. The proposed function-space predictive objective is jointly optimized with the segmentation objective, while the auxiliary predictive components are removed during inference.
All models are trained for $300$ epochs on a server equipped with eight NVIDIA RTX $4090$ GPUs. Main benchmark comparisons are averaged over three independent random seeds. Controlled ablation and cross-backbone studies use two shared seeds for all compared configurations
under identical training and evaluation protocols. 

\textbf{Comparison with State-of-the-Art Methods:}
Table~\ref{tab:main-results} reports the comparison between the proposed method and representative medical image segmentation approaches across five benchmarks, including ultrasound, endoscopy, and histopathology datasets. The compared methods include CNN-based architectures, such as U-Net \cite{ronneberger2015unet}, UNet++ \cite{zhou2018unetpp}, UNeXt \cite{valanarasu2022unext}, Rolling-UNet-S \cite{liu2024rolling} and PBE-UNet \cite{wang2026pbe}, as well as recent KAN-based segmentation models, including U-KAN \cite{li2025ukan}, MM-UKAN++ \cite{MMUKANpp}, KMUNet \cite{zhang2025kmunet}, Implicit U-KAN2.0 \cite{cheng2025implicit} and UUEKAN \cite{CHEN2026110649}.

Our method consistently achieves superior performance across most
datasets, obtaining the highest average Dice and IoU scores of $83.37\%$ and $75.04\%$, respectively. Compared with the matched scaffold, FS-JEPA improves the average Dice from $82.52\%$ to $83.37\%$, corresponding to gains of $0.85$ percentage points, respectively. Since the two models share the same segmentation architecture and inference-time parameter count, these improvements isolate the contribution of the proposed function-space predictive objective. FS-JEPA also outperforms the strongest competing KAN-based method, UUEKAN, by $+2.25$ percentage points Dice percentage points while using substantially fewer inference-time parameters.
These results indicate that the performance improvement mainly originates from function-space predictive supervision rather than increased model capacity. As shown in Fig.~\ref{fig:sixdataset}, qualitative comparison demonstrates that the proposed framework also achieves competitive performance across different 
imaging modalities, producing more accurate boundary delineation and lesion localization. By learning multi-radius KAN edge-function signatures, the framework provides more effective functional representations for challenging medical segmentation scenarios with complex appearance variations and ambiguous boundaries. Overall, these results demonstrate the effectiveness of predicting 
structured function representations beyond conventional feature embeddings for KAN-based segmentation.

\textbf{Mechanistic Analysis and Cross-Backbone Generalization:}
As shown in Fig.~\ref{fig:mechanistic_analysis}(a), FS-JEPA improves Dice in eight of the nine backbone--dataset settings, with an overall average
gain of $0.47$ percentage points. The gains across U-KAN, UUEKAN, and KMUNet suggest that FS-JEPA is not restricted to a specific segmentation scaffold.
Fig.~\ref{fig:mechanistic_analysis}(b1--b3) shows that compensated
perturbations $(+\delta,-\delta)$ leave the aggregated response and
segmentation loss nearly unchanged, while the FS-JEPA objective remains
sensitive to these edge-level variations. Fig.~\ref{fig:mechanistic_analysis}(c) further shows that FS-JEPA induces edge-specific changes in KAN function responses under paired initialization, edge indices, and input grids. Together, these results indicate that FS-JEPA provides an explicit pre-aggregation learning signal and directly reshapes KAN edge-function representations.

\textbf{Ablation Studies:}
We conduct ablation studies to investigate three aspects of the proposed
function-space predictive learning framework: (1) the effect of different
predictive target spaces, (2) whether multi-radius signatures improve function prediction quality, and (3) how different signature constructions affect the downstream segmentation performance.

\textbf{Effect of Predictive Target Space:}
Table~\ref{tab:target-space-ablation} compares different predictive targets under the same segmentation scaffold and optimization budget. Without predictive learning, the baseline achieves a Dice score of $0.7808$. Token- and Node-JEPA obtain $0.7808$ and $0.7845$, respectively, indicating that feature-level and node-level prediction do not provide consistent benefits.
When prediction is performed on pre-aggregation edge responses, Edge-JEPA
achieves $0.7771$ Dice, suggesting that a single function evaluation
is insufficient to characterize the underlying KAN edge functions. In contrast, the proposed multi-radius edge-function signature achieves the best performance with $0.7926$ Dice. These results demonstrate
that effective function-space predictive learning requires not only moving
beyond feature representations, but also constructing an informative
representation of individual KAN edge functions.

\textbf{Effect of Function Signature Construction:}
Table~\ref{tab:function-signature-diagnostics} evaluates the quality of
function prediction under different signature constructions. Compared with
single-response prediction, Raw multi-point signatures substantially improve function prediction, reducing Relative MAE from $0.9910$ to $0.9175$ and NRMSE from $0.9893$ to $0.8999$. It also increases Pearson correlation from $0.3139$ to $0.4622$, indicating that multi-radius responses better preserve the functional behavior of KAN edges.
Although different normalization strategies improve certain statistics, they do not consistently enhance functional prediction. Raw multi-point achieves the best overall prediction quality, with the lowest regression errors and highest correlation among signature variants.

\textbf{Effect of Signature Design on Segmentation:} 
Table~\ref{tab:function-signature-segmentation} further evaluates the impact of signature construction on the final segmentation task. Compared with Response only, Raw multi-point improves Dice from $0.7771$ to $0.7926$, yielding a gain of $+1.55\%$. Additional normalization strategies do not provide consistent improvements.
Centered and ZCA signatures achieve Dice scores of $0.7786$ and $0.7810$, respectively. Diagonal normalization obtains the highest IoU, but with larger seed variation.
The multi-point signature achieves the highest Dice and is therefore
adopted as the final design.


\section{Conclusion}
\label{sec:conclusion}

We introduced FS-JEPA, a function-space joint-embedding predictive framework that extends predictive learning from feature embeddings to structured KAN edge-function signatures. By predicting multi-radius function signatures before edge aggregation, the proposed method enables direct representation learning over KAN edge functions. The framework is trained jointly with segmentation supervision from random initialization, while all predictive components are removed during inference.
Experiments across medical image segmentation benchmarks demonstrate consistent improvements over existing KAN-based methods, validating the effectiveness of predictive learning in KAN function space.


\bibliography{aaai2027}


\end{document}